\documentclass[letterpaper, 10 pt, conference]{ieeeconf}  

\IEEEoverridecommandlockouts 

\usepackage{cite}
\usepackage{amsmath,amssymb,amsfonts}
\usepackage{algorithmic}
\usepackage{graphicx}
\usepackage{textcomp}
\usepackage{xcolor}
\usepackage{caption}
\usepackage{subcaption}

\def\BibTeX{{\rm B\kern-.05em{\sc i\kern-.025em b}\kern-.08em
    T\kern-.1667em\lower.7ex\hbox{E}\kern-.125emX}}

\graphicspath{{images/}}

\begin{document}

\title{\LARGE \bf Auto-TransRL: Autonomous Composition of Vision Pipelines for Robotic Perception}

\author{Aditya Kapoor, Nijil George, Vartika Sengar, Vighnesh Vatsal and Jayavardhana Gubbi
\thanks{The authors are with TCS Research \& Innovation,
        Tata Consultancy Services, Bengaluru, Karnataka - 560066, India. e-mail:
        {\tt\small aditya.kapoor1, george.nijil, vartika.sengar, vighnesh.vatsal, j.gubbi <@tcs.com>}}
}%

\maketitle

\begin{abstract}
Creating a vision pipeline for different datasets to solve a computer vision task is a complex and time consuming process. Currently, these pipelines are developed with the help of domain experts. Moreover, there is no systematic structure to construct a vision pipeline apart from relying on experience, trial and error or using template-based approaches. As the search space for choosing suitable algorithms for achieving a particular vision task is large, human exploration for finding a good solution requires time and effort. To address the following issues, we propose a dynamic and data-driven way to identify an appropriate set of algorithms that would be fit for building the vision pipeline in order to achieve the goal task. We introduce a Transformer Architecture complemented with Deep Reinforcement Learning to recommend algorithms that can be incorporated at different stages of the vision workflow. This system is both robust and adaptive to dynamic changes in the environment. Experimental results further show that our method also generalizes well to recommend algorithms that have not been used while training and hence alleviates the need of retraining the system on a new set of algorithms introduced during test time.


\end{abstract}

\section{Introduction}
One of the most important parts of solving a vision task is to correctly identify the correct sequence of preprocessing steps and the algorithms that would be most suitable for restoring the image to a format that can be used for achieving the goal task. Preprocessing of images and videos plays a very vital role in the performance of a computer vision pipeline. Inappropriate choices of the preprocessing sequence and algorithms can drastically hamper the performance of the goal task. The preprocessing pipeline can have different arrangements and the number of algorithms to choose from are fairly large in number. As a result, there can exist multiple such algorithmic configurations to choose from. On the other hand, for the same task there can be multiple different workflows (eg: image corrupted by changing exposure and then adding noise can be retrieved by both doing exposure correction followed by denoising and by denoising followed by exposure correction) in different system and environment conditions. With such a diverse set of choices the time, effort and resources needed to build a vision pipeline increases exponentially. In many cases, the data available to construct a vision workflow belongs to a fixed distribution and hence building systems with such a constraint can lead to failures when these systems are deployed in the real world due to various uncertainties. In cases where one needs to optimize over memory, energy and time of the entire pipeline, the right choice of algorithms at different stages of the vision workflow becomes increasingly more difficult and complex. Along with these difficulties due to the fast moving nature of this field, new discoveries are made due to which the pool of algorithms to choose from keeps expanding. On the contrary, comparison of all algorithms based on intuition too, can yield suboptimal solutions. Most of the researchers make pipeline choices based on human intelligence. Human intelligence is often limited and the highest performance of selecting an algorithm cannot be achieved based solely on a researcher’s intuition. Human expert-based design is slow, especially in cases when the image has undergone multiple forms of distortions. Thus, there is a need to automate the processes of design choices so that experts can achieve good results rapidly. In addition to developing a fixed pipeline for vision tasks, this system can be incorporated in the perception stack to choose best algorithms from a set based on the input image and goal task to dynamically construct a workflow for the robot to achieve its goal.

Our main contribution is to leverage the Transformer Architecture \cite{NIPS2017_3f5ee243} along with Deep Reinforcement Learning techniques to search the algorithmic space such that it can generalize well to the set of algorithms that were not used during training. In a nutshell, after the sequence of preprocessing steps are decided, our framework performs a knowledge based graph search over the algorithmic space at every stage of the pipeline and identifies the algorithms that would be well suited to complete the vision pipeline for a given input image. As our framework can retrieve algorithms dynamically, it reduces the level of human intervention for algorithm selection. Apart from this, it has the ability to adapt to unforeseen algorithms that can be introduced at any point in the search space, hence requiring little to no retraining of the framework.

\section{Background}

\subsection{Approach}
The process of identifying the vision pipeline for a particular task can be thought of a two stage process. Firstly, we put together the high-level vision pipeline (tasks like Denoising, Exposure Correction, Classification, Object Detection etc forming a sequence) that will eventually create the vision workflow. This can be thought of a seq2seq decision making problem. Secondly, selecting the appropriate algorithms for every high-level tasks (Denoising: FFDNet, Exposure Correction: Gamma Correction 0.5, Classification: Resnet-50 etc). In this paper we will address the second stage, that is, algorithm selection and propose a framework that can be used to automate this process. Our work does not construct a new system architecture but on the contrary selects one from the existing set of algorithms that would be a good fit to be a part of the vision pipeline.

The key idea is to make algorithmic choices based on the representation power of the algorithms and improve the selection process over the training period with the help of Deep Reinforcement Learning. In this work, we assume that the high-level sequence of the vision pipeline is known and the decisions are to be made for algorithm selection for every stage of the vision pipeline. Further, we do a graph search using the Transformer Architecture over the algorithmic space and update the embedding, key and query networks with Deep Reinforcement Learning framework using PPO \cite{schulman2017ppo} (Proximal Policy Optimization) as the underlying algorithm. 

\subsection{Problem Formulation}
We consider constructing a vision workflow as a sequential decision-making problem that can be modeled as Markov decision processes (MDPs) \cite{mdp}.  An MDP is defined by \( (\mathcal{S},\mathcal{A},\mathcal{P},\mathcal{R},\rho_0,\gamma ) \), where \(\mathcal{S}\) is the state space, image, \(\mathcal{A}\) is the action space. set of all algorithms, \(\mathcal{P}(s_{l+1}|s_l,a_l)\) specifies the state transition probability distribution, image processing step, \(\mathcal{R}(r_l|s_l,a_l)\) specifies the reward distribution, validation accuracy, \(\rho_0(s_0)\) denotes the initial state distribution, distorted image, and \( \gamma \in (0,1]\) denotes a discount factor. At each timestep, the policy selects an action independently according to its state-conditioned policy \( \pi_i(a_l|s_l;\theta) \), where \(s_l\) denotes the state information available to the policy and \( \theta \) denotes its parameters. The policy subsequently earns a reward \(r_l\) sampled from \(\mathcal{R}\), and the environment undergoes a state transition, \(s_{l+1} \sim \mathcal{P}(\cdot|s_l,a_l)\).

\begin{figure}[!ht]
        \centering
        \includegraphics[width= \linewidth]{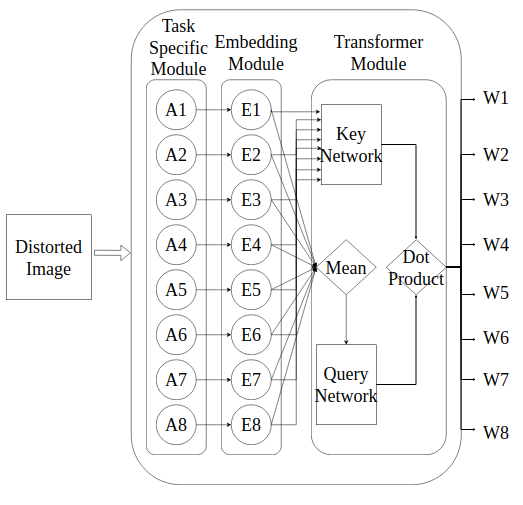}
        \caption{\textbf{Policy Architecture}
        }
        \label{fig:policy}
\end{figure}

We focus on efficiently solving the \textit{algorithm selection} task, wherein at each timestep the pipeline progresses one step further, and the policy (see Fig \ref{fig:policy}) attempts to maximize the rewards. More precisely, we wish to find the optimal policy parameters that solve \( \theta^* = \underset{\theta}{\mathrm{argmax}}\ J(\theta) \), where
    \begin{equation}
        \label{eq:obj}
        J(\theta) = \mathbb{E} \Big[ \sum_{l=0}^L \gamma^l r_l \Big],
    \end{equation}
    
    \noindent with \(L\) denoting the length of the vision pipeline.  By the policy gradient theorem \cite{pg}, the gradient of \(J\) with respect to the policy parameters \(\theta\) is given by
    \vspace{-.1cm}
    \begin{multline}
        \label{eq:pg}
        \nabla_{\theta} J(\theta) = \mathbb{E} \Big[ \nabla_{\theta} \log \pi(a_l|s_l)  \\ \left(Q^{\pi}(s_l,a_l)- b(s_l,a_l) \right) \Big],
    \end{multline}

    \noindent where \(Q^{\pi}(s_l,a_l)\) denotes the expected future reward, and \(b(s_l,a_l)\) is commonly known as a \textit{baseline function}, which can be any any function that depends on the state, and the actions at length \(l\). Often a learned value function is used for the baseline but in our case, we use a running mean of the validation accuracy from previous episodes for the baseline.

\section{Method}
During training, we distort the image in either of the 4 sequences; change exposure then add noise, add noise change exposure, change exposure only and add noise only. For all of the distortion sequences, we fix the high-level pipeline (denoising followed by exposure correction, exposure correction followed by denoising, exposure correction and denoising respectively for each of the 4 sequences of distortion) such that the distorted image is retrievable to its original format. We call our approach \textit{Auto-TransRL}. In order to aid our search algorithm, we train individual Artificial Neural Networks, for each attribute level of the image (exposure: under/well/over exposed; noise: no/low/mid/high level noise). By doing so we construct a knowledge base that enables our framework to perform a guided search over algorithm set. For example, if the State Attribute Identifier for exposure detected that the input image is underexposed, the search algorithm is restricted to perform its search over algorithms that are eligible for correcting underexposed images.

\subsection{State Attribute Identifiers (SAI) a.k.a Knowledge base}
We train 2 different neural networks to identify exposure and noise levels of the input image. We have bifurcated the noise levels into 4 categories, that is, no-noise-level, low-noise-level, mid-noise-level and high-noise-level and exposure levels into 3 categories, that is, under-exposed, well-exposed and over-exposed. The State identifiers complement the search algorithm to restrict the search over the set of algorithms that would be suited to address the distortion the input image has undergone. In this way, as we introduce domain knowledge while performing the search, we ensure that convergence is achieved at a faster rate.
The \textit{SAI} Neural Networks are trained in a Supervised way wherein they need to classify the distortion level in the input image. We use the CIFAR-10 \cite{CIFAR10} Image dataset and distort the images and associate these distortion levels with their respective labels. We use Resnet-50 as the backbone of both the state identifiers. There are other ways to extract the state attribute levels of an image as the ones used by \cite{sengar2021automatic} but they require manual intervention at various points in the decision making process. 

\subsection{Auto-TransRL}
Our proposed approach connects a sequence of \textbf{Policies} according to the high-level preprocessing sequence. Every policy has 3 modules namely \textbf{Task Specific Module ‘TSM’}, \textbf{Embedding Module ‘EM’} and \textbf{Transformer Module ‘TM'}. Every policy's TSM has a set of algorithms that serve a very specific purpose such as “Edge Detection”, “Classification”, “Exposure Correction” etc. Each TSM is followed by an EM. Each algorithm in the TSM is associated with an embedding network in the EM. The EM ensures that the output of all algorithms are mapped to the same output dimension. The EM is further followed by a TM. The TM consists of a Key network, ‘K’ and Query network, ‘Q’. The score generated by the dot product between the key and query vectors is further used to calculate the weights corresponding to each algorithm in TSM. 
We train individual policies to select an algorithm that achieves a specific task in the vision pipeline. Thus, in our use case, we train 3 such policies for Exposure Correction, Denoising and Classification. The weights produced by the TM are used as the policy output. 

In our use case the embedding networks in EM, in each policy, shares the weight parameters (not a necessary condition). The embedding networks are non-linear MLPs (Multi Layer Perceptrons). 
The Key and Query Networks in the Transformer Module are MLPs too. The Query network takes as input a mean of all the algorithm (in the TSM) embeddings to generate a global query vector which after dot product with the key vectors outputs a relative weight corresponding to every algorithm in the TSM.

This entire setup is making decisions based on the representation power of every algorithm in the TSM. As the weights generated by the TM is the measure of a similarity score between the mean of every algorithm's output against each algorithm's output, it acts as a good metric to select an algorithm. Hence higher the weight the better an algorithm is on average because the weight values are a direct measure of an algorithm’s representation power.

We train every policy to choose an algorithm for a specific task using PPO and the classification accuracy is used as the reward signal for all the policies. Again, for the classification task, we use the CIFAR-10 image dataset.

\begin{figure}[!ht]
        \centering
        \includegraphics[width= \linewidth]{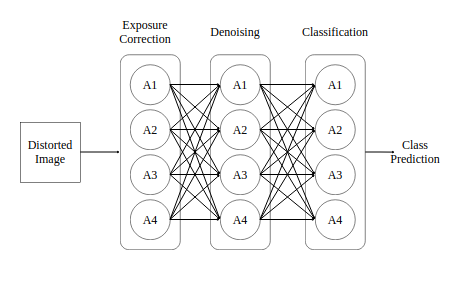}
        \caption{\textbf{Zoomed out perspective of Auto-TransRL.}
        }
        \label{fig:meta-neural-network}
\end{figure}

\subsection{Background}
Within the policy the only things learned are the networks in the Embedding Modules and Transformer Module. The algorithms in Task Specific Module are pre-trained and are not updated during the training process. As a result, the algorithms in the TSM convert an input image to a latent embedding that belongs to a fixed and learned distribution. Thus the EM in conjunction with TM learns to choose algorithms solely based on the representation power of every algorithm in the TSM. We believe that the latent embedding generated by each algorithm captures the information about the distortions that have been made on the input image. This assumption is based on the empirical evidence that the performance of algorithms suffer if the image is distorted in any manner. For example, the classification accuracy for a particular pretrained model for a distorted image dataset would be less when compared to one with no distortions. 

From a zoomed-out perspective, one can think of the Policy in the vision workflow as a layer of a meta-neural network, (refer Fig \ref{fig:meta-neural-network}) wherein each individual neuron in a layer is a complex function, a.k.a neural network itself and the idea of knowledge guided search acts as a “thoughtful” dropout layer. Also every layer specializes in a very specific task and this can be correlated to the CNN layers in a classifier. Many such concepts from Artificial Neural Networks can be extended to our proposed approach. We haven’t explored this direction actively but we hope to do so in the near future.

\begin{figure}[h!]
\subfloat[Classification Accuracy when known algorithms are used in TSM during test time\label{fig:test1}]
  {\includegraphics[width=0.9\linewidth]{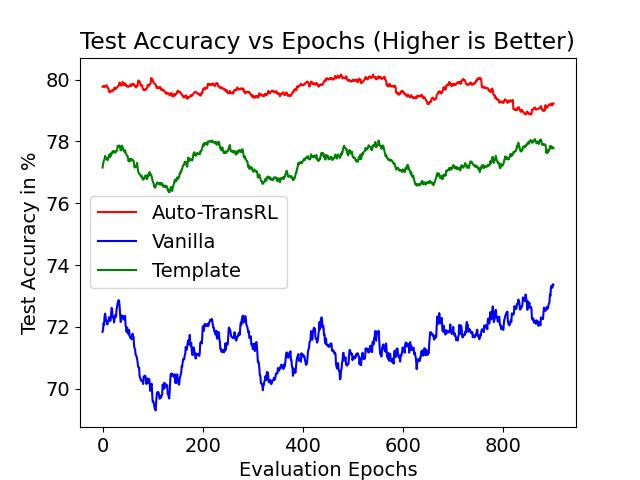}}\hfill
\subfloat[Classification Accuracy when a mixture of both known and unknown algorithms are used in TSM during test time\label{fig:test2}]
  {\includegraphics[width=0.9\linewidth]{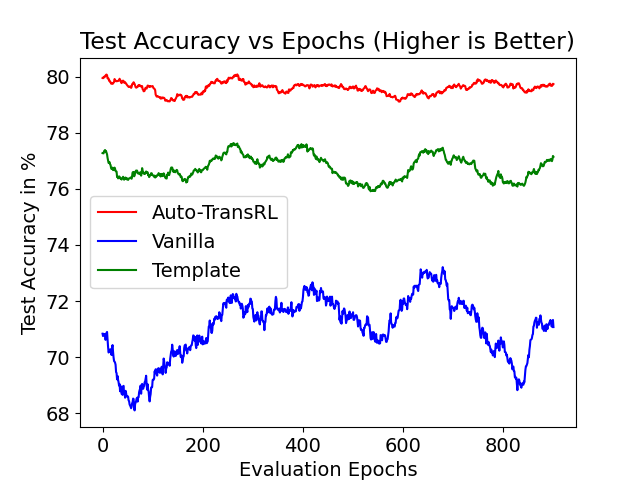}}\hfill
\subfloat[Classification Accuracy when unknown algorithms are used in TSM during test time\label{fig:test3}]
  {\includegraphics[width=0.9\linewidth]{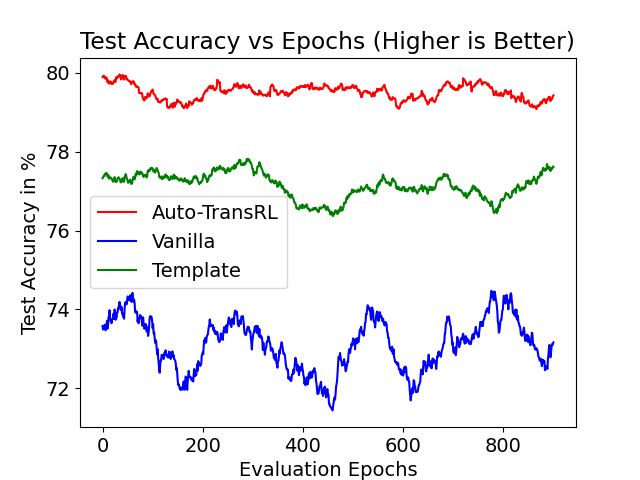}}
\caption{Classification Task on different distorted CIFAR-10 images}
\end{figure}

\section{Experimental Results}
We experimentally evaluate the performance of Auto-TransRL on classification, exposure correction and denoising tasks. We compare average episode reward against various image distortions between Auto-TransRL, Template based approaches and by directly feeding into a classifier without any preprocessing layers (Vanilla). To validate our claim that Auto-TransRL generalizes well to unseen algorithms, we compare the performance of Auto-TransRL and the other baselines on partially-known and unknown settings. In the former experimental configuration, we add 4 unseen algorithms along with the ones used during train time and in the latter we use only unseen algorithms. 

\subsection{Comparison to Template-based and Vanilla approaches}
To evaluate the effect of Auto-TransRL on adapting tendency and
performance, we compare our Auto-TransRL algorithm to the following baseline approaches on a set of vision tasks:
\begin{itemize}
    \item \textbf{Template Based Systems (Template)}: For a given distorted image we fix the algorithm in order to restore it back to its original format.
    \item \textbf{Vanilla System (Vanilla)}: For a given distorted image we directly pass it through a classifier.
\end{itemize}

We considered Vanilla System as a baseline because it represents the naive approach of dealing with the actual task without considering any preprocessing layers. Vanilla Systems are the most easily deployed baselines for any vision task. One would make use of it when the classifier in use has been trained on similar distorted images. We additionally considered Template Based System as a baseline because it represents the current way of dealing with perception modules in robot systems, that may work well for many problems in which there is little to no deviation from the training setup. We hypothesize that Auto-TransRL will selectively consider only the the algorithms that will allow the entire system to achieve higher classification accuracy. Intuitively, for each image, Auto-TransRL attempts to find the best algorithm or no algorithm (when the image is not distorted) from the set of algorithms that would fit in the pipeline to restore the image and classify its type.

We evaluate Auto-TransRL, Template and Vanilla in three task settings, which we refer to as \textit{Known, Partially-Known} and \textit{Unknown} test beds. 

\textbf{Known :} In this setting, the \textbf{TSM}, during test time, comprises of only algorithms that were used during training. 

\textbf{Partially-Known :} In this setting, the \textbf{TSM}, during test time, comprises of a mixture of both algorithms that were used during training and the ones that were not.

\textbf{Unknown :} In this setting, the \textbf{TSM}, during test time, comprises of only algorithms that were not used during training.

We compare Auto-TransRL, Template based System, and Vanilla System in all the three tasks. As predicted, Vanilla System fails to make significant progress towards solving the tasks (Fig. \ref{fig:test1}, Fig \ref{fig:test2}, Fig \ref{fig:test3}).  Auto-TransRL outperforms the Template based System, achieving a higher classification accuracy consistently.

\section{Conclusion}
In this paper, the proposed method is a search-based framework which chooses algorithms based on their representation capabilties. This allows it to be more generic for decision problems. As a result, this will enable Perception systems in Robots to make use of the pre-existing algorithms to quickly and dynamically configure a perception pipeline for a task in different environment and system configurations. Additionally, we also demonstrate that our algorithm, Auto-TransRL can adopt to new algorithms that were not used during train time and still recommend optimal pipelines. Auto-TransRL can also be extended to compare performance of algorithms in a data driven way. In future, we would demonstrate, on a robot test bed, how Auto-TransRL can be used to dynamically build a Perception pipeline. We are also looking at different ways to train all the policies together with a new dense reward function to tackle the Credit Assignment Problem which we encountered in this work.    


\bibliographystyle{IEEEtran}
\bibliography{IEEEabrv,refs}

\end{document}